\documentclass{article}
\usepackage{microtype}
\usepackage{graphicx}
\usepackage{subfigure}
\usepackage{booktabs}
\usepackage{xcolor}

\usepackage{hyperref}

\usepackage[accepted]{icml2020}

\icmltitlerunning{Where is the Information in a Deep Neural Network?}

\usepackage[utf8]{inputenc}
\usepackage{amsmath}
\usepackage{amsfonts}
\usepackage{amsthm}
\usepackage{amsxtra}
\usepackage{amssymb}
\usepackage{soul}
\usepackage{hyperref}
\usepackage[nameinlink]{cleveref}
\usepackage{graphicx}
\usepackage{xspace}
\usepackage{enumerate}
\usepackage{natbib}

\newcommand{\F}{\mathcal{F}}
\newcommand{\E}{\mathbb{E}}
\newcommand{\D}{\mathcal{D}}
\newcommand{\KL}[2]{\operatorname{KL}(\,{\textstyle#1}\,\|\,{\textstyle#2}\,)}
\DeclareMathOperator{\tr}{tr}

\newcommand{\w}{\mathbf{w}}

\newcommand{\cut}[1]{{#1}}

\newcommand{\eff}{\text{eff}}

\DeclareMathOperator*{\argmin}{argmin}

\newcommand{\bra}[1]{\left[#1\right]}
\newcommand{\norm}[1]{\left\|#1\right\|}

\newtheorem{thm}{Theorem}[section]
\newtheorem{proposition}[thm]{Proposition}
\newtheorem{lemma}[thm]{Lemma}

\newtheorem{definition}[thm]{Definition}
\theoremstyle{definition}

\newtheorem{rmk}[thm]{Remark}

\def\eg{e.g.\@\xspace}
\def\ie{i.e.\@\xspace}

\begin{document}

\twocolumn[

\icmltitle{Where is the Information in a Deep Neural Network?}

\icmlsetsymbol{equal}{*}

\begin{icmlauthorlist}
\icmlauthor{Alessandro Achille}{amazon,ucla}
\icmlauthor{Giovanni Paolini}{amazon,sns}
\icmlauthor{Stefano Soatto}{amazon,ucla}
\end{icmlauthorlist}

\icmlaffiliation{amazon}{Amazon Web Services}
\icmlaffiliation{ucla}{Department of Computer Science, University of California, Los Angeles}
\icmlaffiliation{sns}{Department of Mathematics, Scuola Normale Superiore, Pisa. Work conducted while the authors were respectively at UCLA and SNS}

\icmlcorrespondingauthor{Alessandro Achille}{aachille@amazon.com}

\icmlkeywords{Information, Deep Learning, Representation Learning, Optimization, Information Bottleneck, Fisher Information}

\vskip 0.3in
]

\printAffiliationsAndNotice{}

\begin{abstract}
Whatever information a deep neural network has gleaned from training data is encoded in its weights. How this information affects the response of the network to future data remains largely an open question.
Indeed, even defining and measuring information entails some subtleties, since a trained network is a deterministic map, so standard information measures can be degenerate. We measure information in a neural network via the optimal trade-off between accuracy of the  response and complexity of the weights, measured by their coding length. Depending on the choice of code, the definition can reduce to standard measures such as Shannon Mutual Information and Fisher Information. However, the more general definition allows us to relate information to generalization and invariance, through a novel notion of  \emph{effective information in the activations} of a deep network. We establish a novel relation between the information in the weights and the effective information in the activations, and use this result to show that models with low (information) complexity not only generalize better, but are bound to learn invariant representations of future inputs. These relations hinge not only on the architecture of the model, but also on how it is trained, highlighting the complex inter-dependency between the class of functions implemented by deep neural networks, the loss function used for training them from finite data, and the inductive bias implicit in the optimization.
\end{abstract}

\section{Introduction}

After training a deep neural network, all that is left of past experience is a set of values stored in its parameters, or {\em weights.} So, studying what ``information’’ they contain seems a natural starting point to understand how deep networks learn. But how is the information in a deep neural network even defined? Once trained, the weights of a neural network are fixed, and the output is a deterministic function of its input, which has degenerate (infinite) Shannon Mutual Information with the input.
Alternate measures based on ideas from Fisher or Kolmogorov can be used, but they do not relate directly to relevant concepts such as generalization (Fisher), or they cannot be measured in practice for modern deep neural networks (Kolmogorov).
To the best of our knowledge, this paper is the first to connect information quantities that can be measured  even for large neural networks, such as the trace of the Fisher Information Matrix, to desirable properties of the representation of unseen data. These properties include generalization, which depends on Shannon's information, and sensitivity to nuisance variability, which depends on Fisher's.

With few exceptions, reviewed in the next section, information-theoretic approaches to deep learning focus on the \emph{activations} of the network --  the output of its layers -- rather than on the weights. The weights are a representation of {\em past} data (the training set of inputs and outputs), trained to infer statistics of the training set itself (\eg, the outputs). But we do not care to guarantee properties of these inferences, since they pertain to data that we will never see again.
Instead, we wish to guarantee properties of the network in response to {\em future} input data, the so-called test set. Such a response is encoded in the activations, which are representations of unseen data. These representations should ideally be maximally informative ({\em sufficient}) of future outputs in response to future inputs, and minimally sensitive ({\em invariant}) to variability that is irrelevant to the task ({\em nuisances}). Unfortunately, we have no access to future data, so the key question we tackle is: {\em how are desirable properties such as sufficiency and invariance of the representation of future data, related to information quantities measurable from past data, consisting of a finite training set?}

\cut{Sufficiency alone is trivial --- any invertible function of future input data is sufficient -- but comes at the expense of complexity of the representation, and therefore invariance as we articulate in \Cref{sec:induced}. Invariance alone is similarly trivial -- any constant function is invariant. A learning criterion therefore trades off accuracy, complexity and invariance. We define information via \emph{their best achievable  trade-off.} The challenge is that we wish to characterize sufficiency and invariance of representations of the \emph{test} data, while only having access to the training set.}

\subsection{Summary of Contributions and Related Work}
\label{sec:related}

This paper addresses four distinct concepts: (1) Sufficiency of the weights, measured by the training loss; (2) minimality of the weights, measured by the information they contain; (3) sufficiency of the activations, captured by the test loss which we cannot compute, but that we can bound \cut{ using the Information in the Weights}; (4) minimality and invariance of the activations, \cut{which are properties of test data} that are not explicit when training a deep neural network (DNN), but that \cut{we show to} emerge from the learning process. We formally define \textit{information of the weights} and introduce the \cut{notion of} effective information in the \textit{activations}. This is our {\bf first contribution}: We characterize the Information in the Weights of a DNN as the trade-off between the perturbation we \emph{could} add to the weights, and the change in performance the network would exhibit on the task at hand.\cut{ This is the number of bits needed to encode the weights in order to solve the task at some level of precision, as customary in Rate-Distortion Theory.} The optimal trade-off traces a curve that depends on the task and the architecture, and solutions along the curve can be found by optimizing an \emph{Information Lagrangian}. The Information Lagrangian is \cut{in the general form of} an Information Bottleneck (IB) \citep{tishby2000information}, but fundamentally different from the IB used in most prior work \cut{ in Deep Learning},  which refers to the activations rather than the weights
\citep{tishby2015deep,shwartz2017opening,alemi2016deep,higgins17beta,saxe2018information}.
\cut{ Our measure of information is computable even for large networks, and depends on the finite number of training samples.} The fact that some weights can be perturbed at little loss has been known and used for decades \citep{hinton1993keeping, hochreiter1997flat,kingma2015variational}. Our definition starts from this observation and encompasses classical notions of information such as  Shannon's and Fisher's. \cut{Our paper develops and expands the formalism introduced by \citet{achille2019information} to further  the connections between the optimization algorithm used to train a network, the geometry of the loss landscape, and properties of the activations.}

Our {\bf second contribution} is to derive a relation between the information in the weights, which are functions of past data, and the information in the activations, which are functions of unseen data (\Cref{sec:re-emergence}). We show that the Information Lagrangian (IL) bounds the Information Bottleneck of the activations (IB), but not vice-versa. This is important, as the IB of the activations is degenerate when computed on the training set (\cite{saxe2018information}) and cannot be used to enforce any desirable properties at training time. On the other hand, the IL remains well defined, and through our bounds it controls invariance at test time. Our result tightens and extends the Emergence Bound introduced by \cite{achille2018emergence}, making it computable for finite datasets.

Our definition of information reduces to well-known ones as special cases, including  Fisher's and Shannon's. Unfortunately, neither alone captures desirable properties of the representation of test data while being computable on a finite dataset: Shannon's  relates to generalization via the PAC-Bayes Bound (\Cref{sec:pac-bayes}), but cannot be computed using a single dataset. Fisher's can be computed, and we show it to be related to invariance (\Cref{sec:re-emergence}), but does not directly inform generalization. Our {\bf third contribution} is to characterize the connection between the two, thus establishing a link between invariance and  generalization. Although it is possible to minimize Fisher or Shannon Information independently, we show that when the weights are learned using a stochastic optimization method, the two are bound together to first-order approximation. This result is made possible by the flexibility of our framework (\Cref{sec:sgd-dynamics}). Finally, in \Cref{sec:discussion} we discuss open problems and further relations with prior work.

\noindent{\bf Important caveats:} Measuring the information in the weights requires a choice of encoding, represented by an arbitrary ``pre-distribution'' and ``post-distribution.'' This can be confusing at first reading, and the freedom may appear unwarranted. On the contrary, this is essential to establishing the relation between computable quantities and desirable properties of the trained network. It is important to never confuse the pre- and post-distributions, which are arbitrary choices in defining an encoding, with the prior and posterior distribution of the weights in Bayesian Deep Learning: Once trained, the weights are fixed, so the posterior is degenerate, and the pre-distribution is not a prior in the Bayesian sense. To avert any possible confusion, we use the terms pre- and post- instead of prior and posterior. It is also important to not confuse the post-distribution -- which can be chosen regardless of the optimization algorithm  -- with the ensemble distribution observed {\em during training}, due for instance to using stochastic gradient descent or any of its variants, like Langevin Dynamics, which we refer to collectively as SGD. The posterior is not free for us to choose, and becomes trivial after training ends. The post-distribution, one the other hand, is arbitrary and well defined after training. \cut{If a posterior is available for whatever reason, we are free to choose the post-distribution to coincide with it, but we are not bound to this choice, as the post-distribution is arbitrary.} The particular choice of post-distribution that coincides with the distribution induced by SGD  yields a measure of information that is neither Shannon's nor Fisher's, but sheds light on the interplay between them, the dynamics of deep learning, and the properties of the trained network.

\noindent{\bf Related works} on information properties of the weights \citep{russo2019much,xu2017information,pensia2018generalization,negrea2019information,asadi2018chaining} bound the generalization gap of a training algorithm using the Shannon Information in the weights and a notion of ``information stability''. This is orthogonal and complementary to our work, as we do not aim to establish yet another generalization bound. One of our results (\Cref{prop:weight-shannon-fisher-relation}) relates flat minima (low Fisher Information) to path stability of SGD \citep{hardt2015train} and information stability \citep{xu2017information}, showing that optimization algorithms that converge to flat minima and are path stable also satisfy a form of information stability.  We show that, in realistic settings for a DNN, if the optimization algorithm is stable \cut{(in the sense that changes in the final point of the optimization are small relative to perturbations of the dataset)} then, together with the minimization of the Fisher, this implies information stability, and hence generalization, by either the PAC-Bayes bound, which we use, or the bound proposed by \citep{xu2017information}. While the relaxed loss used by \citep{xu2017information} is related to ours, they use Gibbs' algorithm to minimize it, which is not practical for a real DNN. We show that more practical variants of SGD  minimize a free energy, thus connecting the theory with common practice and emphasizing the role of the dynamics of the training process to obtain generalization as never done before. Unlike those works, ours does not depend solely on the noise induced by SGD, but also on the geometry of the loss landscape, which depends on the dataset, the loss function and the architecture.
Moreover, we connect information stability not only to generalization, but also to the information in the activations and the resulting invariance properties (\Cref{sec:info-activations}). No existing work we are aware of addresses the fact that the information in a trained network must depend on the finite training set, the architecture, the optimization, and must be actually computable without knowledge of the distribution from which the data is drawn.  \cut{Also related to our work is \citet{goldfeld2018estimating}, who estimate mutual information under the hypothesis of inputs with isotropic noise.}

\subsection{Preliminaries and Notation}
\label{sec:notation}
We denote with $x \in X$ an input (\eg, an image), and with  $y \in Y$ a ``task variable,'' a random variable which we are trying to infer, \eg, a label $y \in Y = \{1, \dots, C\}$. A dataset is a finite collection of samples $\D = \{(x_i, y_i)\}_{i=1}^N$ that {\em specify} the task.
A DNN model trained with the cross-entropy loss encodes a conditional distribution $p_w(y|x)$, parametrized by the weights $w$, meant to approximate the posterior of the task variable output $y$ given the input $x$.
The Kullbach-Leibler, or \textit{KL-divergence,} is the relative entropy between $p(x)$ and $q(x)$: $\KL{p(x)}{q(x)} := \E_{x\sim p(x)}\big[\log (p(x)/q(x))\big]$. It is always non-negative, and zero if and only if $p(x) = q(x)$. \cut{It  measures the (asymmetric) similarity between two distributions.}Given a family of conditional distributions $p_w(y|x)$ parametrized by a vector $w$, we can ask how much perturbing the parameter $w$ by a small amount $\delta w$ will change the distribution, as measured by the KL-divergence. To second-order, this is given by $\E_x\KL{p_w(y|x)}{p_{w+\delta w}(y|x)} = \delta w^t F \delta w + O(\|\delta w\|^3)$
where $F$ is the \textit{Fisher Information Matrix} (or simply ``Fisher''), defined by $
F := \E_{x, y\sim p(x)p_w(y|x)}[\nabla \log p_w(y|x)^t\nabla \log p_w(y|x)] = \E_{x\sim p(x)p_w(y|x)}[- \nabla^2_w \log p_w(y|x)]$
\citet{martens2014new}. Note that the Fisher depends on the ground-truth data distribution $p(x,y)$ only through the domain variable $x$, not the \emph{task variable} $y$, since $y\sim p_w(y|x)$ is sampled from the model distribution, not the real one, when computing the Fisher. This property will be used later.
Given two random variables $x$ and $z$, their \textit{Shannon mutual information}  is defined as $I(x; z) := \E_{x \sim p(x)}[\KL{p(z|x)}{p(z)}]$
that is, the expected divergence between the prior distribution $p(z)$ of $z$ and the distribution $p(z|x)$ after an observation of $x$. It is positive, symmetric, zero if and only if the variables are independent \citep{cover2012elements}.
In supervised classification, one is usually interested in finding weights $w$ that minimize the cross-entropy loss $L_\D(w) := \sum_{(x,y) \sim \D}[-\log p_w(y|x)]$ on the training set $\D$. The loss $L_\D(w)$ is usually minimized using \textit{stochastic gradient descent} (SGD) \cite{bottou2018optimization}, which updates  the weights $w$ with an estimate of the gradient computed from a small number of samples (mini-batch).
That is, $w_{k+1} = w_k - \eta \nabla \hat{L}_{\xi_k}(w)$,
where $\xi_k$ are the indices of a randomly sampled mini-batch and
$\hat{L}_{\xi_k}(w) = \frac{1}{|\xi_k|} \sum_{i \in \xi_k} [-\log p_w(y_i|x_i)]$. Notice that $\E_{\xi_k}[\nabla \hat{L}_{\xi_k}(w)] = \nabla L_\D(w)$, so we can think of the mini-batch gradient $\nabla \hat{L}_{\xi_k}(w)$ as a ``noisy version'' of the real gradient.
Using this interpretation, we write:
\begin{equation}
\label{eq:sgd-diffusion}
w_{k+1} = w_k - \eta \nabla L_\D(w_k) + \sqrt{\eta}\  T_{\xi_k}(w_k)
\end{equation}
with the induced ``noise’’ term $T_{\epsilon_k}(w) = \sqrt{\eta}\,\big(\nabla \hat{L}_{\xi_k}(w) - \nabla L_\D(w)\big)$. Written in this form, \cref{eq:sgd-diffusion} is a Langevin diffusion process with (non-isotropic) noise $T_{\xi_k}$ \citep{li2015stochastic}.
We say that a random variable $n$ is a {\em nuisance} for the task $y$ if $n$ affects the input $x$ but is not informative of $y$, \ie $I(n, y) = 0$. We say that a representation $z$ is maximally invariant to $n$ if $I(z, n)$ is minimal among all sufficient representations, which are all\cut{\footnote{There are infinitely many, for instance the trivial function $z = x$, and any invertible function of it.}} the representations $z$ that capture all the information about the task contained in the input, $I(z, y) = I(x, y)$.
\cut{To avert the risk of any confusion, we emphasize that the stochasticity introduced by SGD or any of its variants, which yields a posterior distribution of weights in independent training trials, has nothing to do with the choice of code, or pre-distribution, and the resulting coding length, or post-distribution, in the definition of Information in the Weights, which we introduce next. The former is anisotropic, non-Gaussian and largely out of our control. The latter is an arbitrary choice akin to a choice of unit in measuring information.}

\section{Information in the Weights}
\label{sec:information}

We could define the Information in the Weights as their coding length after training, but it would not be meaningful as only a small subset of the weights matter in a DNN. Imagine applying a perturbation $w' \gets w + \delta w$ \cut{to the weights $w$,} and observing no change in the cross-entropy loss $L_\D(p_{w'}) \approx L_\D(p_w)$. One would say that such weights contain ``no information'' about the task. Storing them with low precision, pruning, or randomizing them would have no effect on input-output behavior. Conversely, if slight changes were to yield a large increase in loss, one could say that such weights are very ``informative'' and store them with high precision. In other words, we measure information in a weight by how its perturbation affects the loss. But what perturbations should one consider (\eg, additive or multiplicative)? And how ``small'' should they be? What distribution should the perturbations be drawn from? To frame these questions, we introduce the following definition, which requires an arbitrary choice of code, represented by a ``pre-distribution'' $P$, chosen before seeing the dataset, and a measure of coding length, represented by a ``post-distribution'' $Q$, chosen after seeing the dataset. Corresponding to different pre- and post- distribution, we will have different information measures, possibly exhibiting different properties.
\begin{definition}[Information in the Weight (IW)]
\label{def:complexity-dnn}
Consider a finite dataset $\D$, an arbitrary post-distribution $Q(w|\D)$ and an arbitrary pre-distribution $P(w)$.  The complexity of the task $\D$ at level $\beta$ relative to the choice of pre- and post-distributions, is
    \begin{align}
    \label{eq:kl-complexity}
        C_\beta(\D; P, Q) &=
        \E_{w \sim Q(w|\D)}[L_\D(p_w(y|x))] \nonumber \\         &
        + \beta  \underbrace{\KL{Q(w|\D)}{P(w)}}_\text{Information in the Weights},
    \end{align}
    where $\E_{w \sim Q(w|\D)}[L_\D(p_w(y|x))]$ is the loss under the distribution $Q(w|\D)$;
    $\KL{Q(w|\D)}{P(w)}$ measures the entropy of $Q(w|\D)$ relative to $P(w)$.
    If $Q^*(w|\D)$ minimizes (\ref{eq:kl-complexity}) for a given $\beta$, we call $\KL{Q^*(w|\D)}{P(w)}$ the \emph{amount of Information in the Weights} for the task $\D$ at level $\beta$.
\end{definition}
The subscript $\D$ reminds us that the loss $L_\D$ is computed on the finite training set, so the IW depends on the number of samples $N$ and does not require knowledge of the data distribution. We call $Q(w|\D)$ post-distribution to clearly distinguish it from the posterior \cut{distribution of the weights} $P(w|\D)$ in Bayesian deep learning, which is entirely different from our approach as articulated in \Cref{sec:related}. Similarly, $P(w)$ is an arbitrary pre-distribution, distinct from a Bayesian prior. Our definition is compatible with a deterministic training process, with a stochastic training process that yields a single set of weights, or with a process that yields a posterior distribution of them. A special case of \cut{the complexity measure in} \cref{eq:kl-complexity} is $\beta=1$, when it formally coincides with the evidence lower-bound (ELBO) used to train Bayesian Neural Networks, but while the ELBO assumes the existence of a Bayesian posterior $P(w|\D)$, which $Q(w|\D)$ approximates, we make no such assumption. \cut{Again, the choice of pre- and post-distribution in our framework is entirely arbitrary, and not bound by a Bayesian construct.}
The ELBO for different \cut{values of} $\beta$ and its connection with rate-distortion theory has also been explored in \citet{hu2018beta}. Closer to our viewpoint is \citet{hinton1993keeping} who show that, for $\beta=1$, \cref{eq:kl-complexity} is the cost to encode the labels in $\D$ together with the weights\cut{ of the network}. This justifies interpreting\cut{, for any choice of $P$ and $Q$,} the term $\KL{Q(w|\D)}{P(w)}$ as the coding length of the weights using some algorithm, although this is true only if they are encoded together with the dataset. These approaches lead to non-trivial results only if $\KL{Q(w|\D)}{P(w)}$ is much smaller than the coding length of the labels in the dataset (\ie, $N \log |Y|$ NATs if we assume a uniform label distribution).\footnote{If that was not the case, we could not exclude that the model is just memorizing all labels in its weights, and hence could not provide a generalization bound using only the description length.} Unfortunately, this is far from reality in typical DNNs, where the description length of millions of weights easily dwarfs the encoding of the labels.  In our framework, what matters is not the particular value of the coding length, but {\em how it changes as a function of $\beta$}, tracing a Pareto-optimal curve which defines the IW. \cut{In the next section, we describe the relation between Information in the Weights and generalization.}

\subsection{Information in the Weights Controls Generalization}
\label{sec:pac-bayes}

\cut{\Cref{eq:kl-complexity} does not immediately relate  to generalization or invariance.}
To relate the information the weights retain about the training set to performance on test data, we start with the well-known PAC-Bayes bound
\begin{thm}[{\citet{mcallester2013pac}, Theorems 2-4}]
\label{thm:pac-bayes}
Assume the dataset $\D=\{(x_i, y_i)\}_{i=1}^N$ is sampled i.i.d.\ from a distribution $p(y,x)$, and \cut{assume that} the per-sample loss used for training is bounded by $L_\text{max} = 1$.\footnote{See \citet{mcallester2013pac} on how to reduce to this case.}
For any fixed $\beta>1/2$, $\delta > 0$, $P(w)$, and  weight distribution $Q(w|\D)$, with probability at least $1-\delta$ over the sample of $\D$, we have:
\begin{align}
\label{eq:pac-bayes-bound}
L_{\text{test}}(Q) &\leq \frac{1}{1-\frac{1}{2\beta}} \Big[\E_{w \sim Q}[L_\D(p_w)] \nonumber\\
&\hfill + \textstyle\frac{\beta}{N} \Big( \KL{Q}{P} + \log \textstyle\frac{1}{\delta} \Big) \Big],
\end{align}
where  $L_\text{test}(Q) := \E_{x,y \sim p(x,y)} [\E_{w\sim Q}[-\log p_w(y|x)]]$ is the expected per-sample test error that the model incurs using the weight distribution $Q(w|\D)$. Moreover, given a distribution $p(\D)$ over datasets, we have the following bound in expectation
\begin{align}
\label{eq:pac-bayes-bound-expectation}
\E_\D [L_{\text{test}}(Q)] &\leq \frac{1}{1-\frac{1}{2\beta}} \Big[\E_\D [\E_{w \sim Q}[L_\D(p_w)]]  \nonumber \\
&+ \textstyle\frac{\beta}{N} \, \E_\D[ \KL{Q}{P}] \Big].
\end{align}
\end{thm}
%\vspace{-.4cm}
Hence,  minimizing the complexity $C_\beta(\D; P, Q)$ can be interpreted as minimizing an upper-bound on the test error, rather than merely minimizing the training error.
\citet{dziugaite2017computing} use PAC-Bayes bounds to compute a non-vacuous generalization bound using a \cut{(non-centered and non-isotropic)} Gaussian prior and a Gaussian posterior.

\subsection{Shannon vs.\ Fisher Information in the Weights}
\label{sec:complexity-measures}

\Cref{def:complexity-dnn} depends on an arbitrary choice of pre- and post-distributions. While this may appear cumbersome, this flexibility allows connecting information quantities computed from the training set to desirable behavior of the representation of test data. In this section, we relate different choices of pre- and post-distributions to well-known definitions, in particular \citep{shannon1948mathematical} and \citep{fisher1925theory}.
In some cases, there may be a true prior $\pi(\D)$ over training sets. Then, a natural choice of pre-distribution $P(w)$ is the one that minimizes the expected test error bound in  \cref{eq:pac-bayes-bound-expectation}, which we call the \emph{adapted prior}. The following proposition relates the measure that minimizes the bound in expectation with the Shannon Mutual Information between weights and dataset.

\begin{proposition}[Shannon IW]
  \label{prop:shannon-complexity-equivalence}
	Assume \cut{the dataset} $\D$ is sampled from \cut{a distribution} $\pi(\D)$, and let the outcome of training \cut{on a sampled dataset $\D$} be described by a distribution $Q(w | \D)$. Then the pre-distribution $P(w)$ minimizing the expected complexity $\E_\D[C_\beta(\D; P, Q)]$ is the marginal $P(w) = \E_\D[Q(w|\D)]$, and the expected Information in the Weights is given by
	\begin{equation}
		\E_\D[\KL{Q}{P}] = I(w; \D)
	\end{equation}
	which is Shannon's mutual information between the weights and the dataset.
	Here, the weights are seen as a (stochastic) function of the dataset, given by the training algorithm.
\end{proposition}
\cut{The proof is straightforward but reported in the Appendix for completeness.} The pre-distribution $P(w)$ is optimal given the choice of the training algorithm (\ie, the map $A: \D \to Q(w|\D)$) and the distribution of training datasets $\pi(\D)$. With this choice, we have the following expression for the expectation over $\D$ of \cref{eq:kl-complexity}:
\begin{align}
\label{eq:IBL-weights}
	\E_\D[C_\beta(\D; P, Q)] &= \E_\D[\E_{w \sim Q}[L_\D(w)]] + \beta I(w; \D).
\end{align}
This expression is the general form of an Information Bottleneck (IB) \citep{tishby2000information}, which has been used in Deep Learning \citep{shwartz2017opening}, but focusing on the activations, which are the bottleneck between the inputs $x$ and the output $y$. Unlike that IB, the {\bf Information Lagrangian (IL)} of \cref{eq:IBL-weights} concerns the weights of the network, which are the bottleneck between the training dataset $\D$ and inference on the future test distribution. Hence, the IL directly relates to the training process and the finite nature of the dataset. \cut{Yet, it can yield bounds on  future performance, first introduced by \citet{achille2018emergence} in a more limited setting that did not allow the flexibility needed to obtain the results in this paper using the Fisher Information.}

While the adapted prior of \Cref{prop:shannon-complexity-equivalence} allows computing an optimal generalization bound, it requires averaging with respect to all possible datasets, which in turn requires knowledge of the distribution $\pi(\D)$ and is, in general, unrealistic. At the other extreme, we can consider an uninformative pre-distribution. \cut{We remind the reader that this is an arbitrary choice and does not imply or require that the actual distributions of the weights, if one exists, be Gaussian or isotropic.}

\begin{lemma}[The Fisher is a semi-definite approximation of the Hessian, {\citet[eq. 6 and Sect. 9.2]{martens2014new}}]
    \label{lemma:hessian-fisher}
    The Hessian $H = \nabla^2_w L_\D(w)$ of the loss function can be decomposed as
    \begin{equation}
    \label{eq:hessian-decomposition}
    H = F + \frac{1}{N} \sum_{(x, y) \in \D} \sum_{j=1}^k \big[ \nabla_z L(y, z)|_{z=f_w(x)}\big]_j H_{[f]_j} ,
    \end{equation}
    where $F$ is the Fisher Information matrix, $z=f_w(x)$ is the output of the network for the input $x$, $L(y,z)=-\sum_{j=1}^k \delta_{y,j} \log(z_j)$ is the cross-entropy loss for the input $x$, and $H_{[f]_j}$ is the Hessian of the $j$-th component of $z$.
    Then, if almost all training samples are predicted correctly,
    as we expect in a well-trained network, we have $\nabla_z L(y_i, z) \approx 0$ and $H\approx F$ to first-order approximation.
\end{lemma}
Our definition reduces to the Fisher Information when one chooses the arbitrary pre-distribution and post-distribution to be Gaussian. This does not mean that the posterior distribution of the weights in repeated training trials is Gaussian, nor isotropic. It is just a choice of encoding that happens to reduce our definition of information to Fisher's.
\begin{proposition}[Fisher IW]
  \label{prop:fisher-complexity-equivalence}
	Let $P(w) \sim N(0, \lambda^2 I)$ and $Q(w|\D) \sim N(w^*, \Sigma)$, centered at any local minimizer $w^*$ of the cross-entropy loss obtained with any optimization algorithm.
	For a sufficiently small $\beta$, the covariance $\Sigma$ that minimizes $C_\beta(\D;P,Q)$ is
		\[
		   \textstyle \Sigma^* = \frac{\beta}{2} \left(H + \frac{\beta}{2\lambda^2}I \right)^{-1}.
		\]

		For $\Sigma=\Sigma^*$, the Information in the Weights is
		\begin{align}
	    \label{eq:kl-equal-fisher}
		&\KL{Q}{P} = \frac{1}{2} \log \left|H + \frac{\beta}{2\lambda^2}\right| + \frac{1}{2} k \log \frac{2\lambda^2}{\beta} -k \nonumber \\
		 & \quad + \frac{1}{2\lambda^2}\left[ \|w^*\|^2 + \tr\left[ \frac{\beta}{2} \left( H + \frac{\beta}{2\lambda^2}I \right)^{-1} \right] \right].
		\end{align}

		If most training samples are predicted correctly, we can estimate \cref{eq:kl-equal-fisher} by substituting $H \approx F$.
\end{proposition}

\begin{figure}
    \centering
    \includegraphics[width=.97\linewidth]{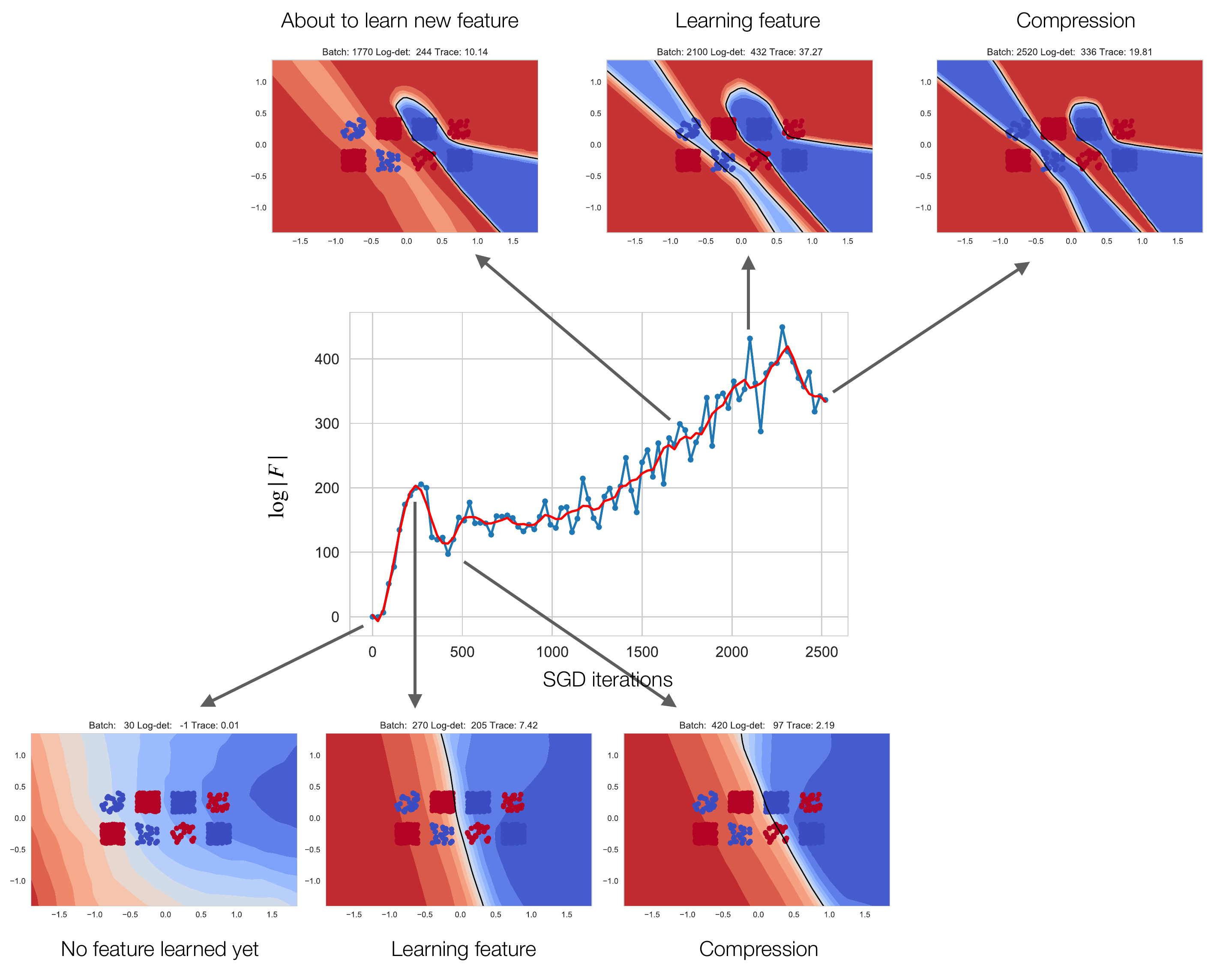}
    \caption{
    Plot of the log-determinant of the Fisher Information Matrix during training of a 3-layers fully connected network on a simple 2D binary classification task. As the network learns an increasingly complex classification boundary, the Fisher increases.
    Moreover, learning of a new feature corresponds to small bumps in the Fisher plot (see \Cref{sec:empirical-verification}).
    }
    \label{fig:bottleneck}
\end{figure}

\begin{rmk}
\label{rmk:hessian-fisher}
The above \cut{proposition} assumes that $w^*$ is a local minimum, otherwise $H$ is not positive definite and the second-order approximation  \Cref{prop:fisher-complexity-equivalence} is invalid, as perturbations along the negative directions can decrease the loss unboundedly. Following \citet{martens2014new}, we use a more robust second-order approximation by ignoring the second part of \cref{eq:hessian-decomposition}, hence using the Fisher as a stable positive semi-definite approximation of the curvature. Then, the approximation \cref{eq:kl-equal-fisher} is valid at all points. The connection between Fisher and Hessian during training has also been studied by \citet{littwin2020convex}.
\end{rmk}

\begin{rmk}[Uninformative prior]
It may be tempting to consider the limit of \cref{eq:kl-equal-fisher} when $P(w)$ becomes an improper prior ($\lambda \to \infty$), in which case the only relevant term is the log-determinant of the Fisher. However, there is an additive term $\log \lambda$ which diverges, as usual when dealing with improper priors. Variational quantities such as the IL in \cref{eq:kl-complexity} may remain well-defined in the limit, as the minimizers are not affected by additive terms independent of $Q(w|\D)$, but see e.g.\ \citet{molchanov2017variational, hron2018variational} for possible pitfalls. However, bounds such as \cref{eq:pac-bayes-bound-expectation} become vacuous and ill-defined in this limit.
\end{rmk}

In \Cref{fig:bottleneck} we plot the log-determinant of the Fisher during training of a simple multi-layer network on a toy binary classification problem, that shows the Fisher growing as the network is learning increasingly complex features, supporting the interpretation of the Fisher as a way of measuring the information contained in the model (see \Cref{sec:experiments} for details). In \Cref{fig:cifar-10} we show a similar result for CIFAR-10.

\subsection{Information in the Learning Dynamics}
\label{sec:sgd-dynamics}

\cut{\Cref{sec:pac-bayes} shows how Shannon Information  controls generalization. In \Cref{sec:re-emergence} we will see that the Fisher Information controls invariance.}
We now return to the question of whether we can just pick one measure of information and use it to characterize both generalization and invariance. In principle, the Fisher  can also be used in \Cref{thm:pac-bayes}, but the resulting bound is likely vacuous as the Fisher is generally much larger than the Shannon Information. However, in this section we argue that, for a deep network trained with stochastic optimization \emph{on a given domain,} Fisher and Shannon are equivalent up to second-order. Informally, \cut{the argument goes as follows:} (i) The Fisher depends on the domain of the data, but not on the labels (\Cref{sec:notation}), hence all tasks sharing the same domain share the same Fisher, (ii) SGD implicitly minimizes the Fisher and, hence, (iii) tends to concentrate solutions in regions of low Fisher, thus indirectly minimizing the Shannon Information given the geometry and topology of the residual loss.
While (i) follows from  the definition, (ii) is not immediate, as SGD does not explicitly minimize the Fisher. The result hinges on the stochasticity of SGD, that tends to preferentially escape sharp minima and, since the Fisher measures the curvature of the loss, tends to evade solutions with high Fisher. To formalize this reasoning, we use a slight reformulation of the Eyring–Kramers law \citep{berglund2011kramers} in \cref{eq:sgd-diffusion}, recently extended to the case of anisotropic noise by \citet{xie2020diffusion}:
\begin{proposition}[{\citet{berglund2011kramers}, eq. 1.9}]
\label{prop:kramers}
Let $w^*$ be a local minimizer of the loss \cut{function} $L_\D(w^*)$. Consider the path $\gamma$ joining $w^*$ with any other minimum with the least increase in loss. The point with the highest loss along the path is a the {\em relevant saddle} $w^s$ with a single negative eigenvalue $\lambda_1(w^s)$.
Then, in the limit of small step size $\eta$,
the expected time for a stochastic gradient-based algorithm with isotropic noise to escape the minimum is given by
\[
\E[\tau] = \frac{2\pi}{|\lambda_1(w^s)|} e^{\frac{1}{T} (\F(w^s) - \F(w^*))},
\]
where the \emph{free energy} is $\F(w) := L_\D(w) + \frac{T}{2}\, \log |F(w)|$, and $T \propto \eta/B$, where $B$ is the batch size. In particular, increasing the temperature parameter $T$  makes the stochastic optimization more likely to avoid minima with high Fisher Information.
\end{proposition}
The proposition above does not apply to SGD, whose noise is highly anisotropic. However, recent work extends this result to anisotropic noise \citep{xie2020diffusion,cheng2019quantitative}. What matters to our thesis is not the numerical value of the escape time, which is different in the isotropic and anisotropic cases, but the tendency of the optimization to converge to flatter vs. sharper minima \cut{as the temperature increases,} which is common to both. We refer the reader to \cite{xie2020diffusion} for specifics on the computation of the rate in the anisotropic case. For our purpose, the proposition above, together with empirical validation by
\citet{hochreiter1997flat,keskar2016large,li2017visualizing,dziugaite2017computing} is sufficient to establish the connection between different information quantities in the rest of this paper. We also remind the reader that we consider any variant of SGD, including Langevin Dynamics, equivalent insofar as they share the bias towards converging to flat minima. Both the empirical evidence and the analysis indicate that the optimization,
rather than minimizing directly the loss function, minimizes a \emph{free energy} \citep{chaudhari2016entropy}  $\F(w) = L_\D(w) + \frac{T}{2}\, \log |F(w)|$, which is connected with the Fisher Information in the Weights.
\cut{The slowing down corresponding to a large Fisher information has also been observed empirically in \citet{achille2018dynamics}.}

\begin{figure}
    \centering
    \includegraphics[width=.6\linewidth]{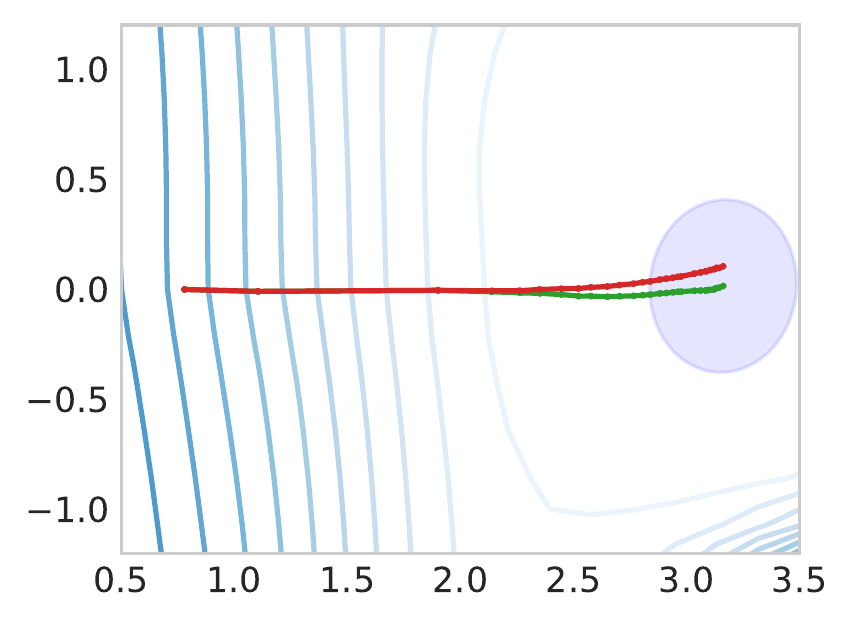}
    \caption{
    \textbf{Stability, Fisher and Information.} For a convolutional network, plot of the training path of two datasets (red and green lines) that differ by one single example, projected onto a plane.
    We used SGD as the training algorithm, with the same random seed and initialization for the two datasets.
    In blue are shown the level curves of the loss function. The amount of Shannon Information that SGD stores in the weights depends on the distance between the final points when training on a perturbed dataset (stability), and the Fisher Information Matrix (whose inverse is shown by the blue ellipse) measures  the local curvature. In this case, the distance between the final points is small relative to the curvature, suggesting that SGD is not memorizing (overfitting) that sample.}
    \label{fig:stability}
\end{figure}

We can now finally address (iii), connecting Fisher and Shannon in DNNs. We leverage a first-order approximation of mutual information from \citet{brunel1998mutual}.
\begin{proposition}[Stochastic optimization links Shannon and Fisher Information]
\label{prop:weight-shannon-fisher-relation}
Assume the space of datasets $\D$ admits a differentiable parametrization.\footnote{
For example by parametrizing the samples through the parameters of a generative models, or more generally using a differentiable sampling method to select the samples of in the dataset.
}
Assume that $p(\D|w)$ is concentrated around a single dataset (\ie, the one used for training), and choose
as post-distribution $p(w|\D) \sim N(w^*, \beta F(w^*)^{-1})$ where $w^* = w^*(\D)$ are the mean weights obtained at the end of training on a dataset $\D$. Then\cut{ we have the approximation}, up to second-order
\begin{align*}
I(w; \D)
&\approx H(\D) - \E_\D\Big[\frac{1}{2} \log\Big(\frac{\beta(2\pi e)^k}{|{\nabla_\D w^*}^t F_w(w^*) \nabla_\D w^*|}\Big)\Big]
\end{align*}
where $H(\D)$ is the entropy of $\D$,
and we assume that $\nabla_\D F(w^*) \ll \nabla_\D w^*$.
\footnote{\cut{That is, that the Fisher does not change much if we perturb the dataset slightly.} This assumption is mainly to keep the expression uncluttered; a similar result can be derived without this additional hypothesis.}
The term $\nabla_\D w^*$ is the Jacobian of the final point with respect to changes of the training set.
\end{proposition}
Again, the Gaussian choice of post-distribution is unrelated to the ``noise'' of SGD: One can pick any optimization method to arrive at a local minimum  $w^*$. Perturbations that are indistinguishable from the input-output behavior of the trained network can be thought of as samples from the distribution $p(w|\D)$, as if one added ``noise'' sampled from $N(0, \beta F(w^*)^{-1})$ to the final solution. This makes it possible to talk about Shannon information, which would otherwise be degenerate for the trained network. Notice that the norm $\|\nabla_\D w^*\|$ of the Jacobian $\nabla_\D w^*$ can be interpreted as a measure of stability of SGD, that is, how the final solution changes if the dataset is perturbed \citep{bousquet2002stability,hardt2015train}.
Hence, reducing the Fisher $F(w^*)$ of the final weights found by SGD (\ie, the flatness of the minimum), or making SGD more stable, \ie, reducing $\nabla_\D w^*$, both reduce the mutual information $I(w;\D)$, and hence improve generalization according to the PAC-Bayes bound.
In the \Cref{fig:stability}, we visualize these quantities computed for a convolutional network. Note that stability or flatness alone may be insufficient to explain generalization: \citet{dinh2017sharp} show examples of sharp but not necessarily stable, minima that generalize.

\section{The Role of Information in the Invariance of the Representation}
\label{sec:re-emergence}
\label{sec:info-activations}

Upon successful training, the {\em weights} of a DNN  are representation of the past that is sufficient (small training loss) and minimal (small information). Sufficiency of the weights implies sufficiency of the {\em activations} (PAC-Bayes), which are functions of future data. What we are missing to complete the program in \Cref{sec:related} is a guarantee that\cut{, in addition to being {\em sufficient},} the activations are also {\em minimal}. One may be skeptical that this is even true: DNNs trained for classification retain enough information on the original image to allow one to reconstruct it from the last layer activations. This is well beyond what is needed to classify, which is just the label, the activations cannot be minimal. {\em Our main contribution is to establish the connection between minimality of the weights and invariance of the activations.}

First, we introduce the notion of \emph{effective information} in the activations: Rather than measuring the information that a hypothetical decoder could extract, we measure the information that the trained network effectively uses to perform the task. We then show that \emph{the Fisher \cut{Information} in the Weights bounds  the effective \cut{Fisher and} Shannon information in the activations}. \cut{Notice that we already related the Fisher Information to the noise of a stochastic optimization algorithm in \Cref{prop:kramers}, or its non-isotropic extensions \citet{xie2020diffusion} for the case of SGD.} Information beyond that needed for the task may still be present, as nothing obliges the network to destroy it, but the trained classifier has no access to it.

\subsection{Induced Stochasticity and Effective Information in the Activations}
\label{sec:induced}

Let $z=f_w(x)$ be the activations of any intermediate layer of a DNN, with $f_w$ a deterministic function. By definition, if information of a particular feature is small or unstable, all perturbed models have equivalent input-output behavior, and therefore the network must not be using it for classification. \cut{Only information that can be extracted from the activations of all perturbed models is effectively used by the network.  This inspires the following}
\begin{definition}(Effective Information in the Activations)
Let $w$ be a weight and $n \sim N(0, \Sigma^*_w)$, with $\Sigma^*_w = \beta F^{-1}(w)$ a perturbation  minimizing \cref{eq:kl-complexity} at level $\beta$ for an uninformative prior (\Cref{prop:fisher-complexity-equivalence}). We call \emph{effective information} (at level $\beta$) the amount of information about $x$ that is retained after the (hypothetical) perturbation:
\begin{equation}
\label{eq:effective-information}
I_{\eff,\beta}(x;z) = I(x;z_n),
\end{equation}
where $z_n = f_{w + n}(x)$ are the activations computed by the perturbed weights $w+n \sim N(w, \Sigma^*_w)$.
\end{definition}
\cut{Again, recall that the choice of Gaussian is just a vehicle for measuring information, and has nothing to do with the noise of SGD nor does it imply that there is an actual distribution of trained weights.} Using this definition, we obtain the following characterization of the information in the activations.
\begin{proposition}
\label{prop:re-emergence}
For small values of $\beta$ we have that
   (i) The Fisher $F_{z|x} = \E_z [\nabla^2_x \log p(z|x)]$ of the activations given the input is:
    \[F_{z|x} = \frac{1}{\beta} \nabla_x f_w \cdot J_f F_w J_f^t\,\nabla_x f_w,\]
    where $\nabla_x f_w(x)$ is the Jacobian \cut{of the representation} with respect to the input, and $J_f(x)$ is the Jacobian with respect to the weights. In particular, the Fisher of the activations goes to zero when the Fisher of the weights $F_w$ goes to zero.
    (ii) If, for any representation $z$, the distribution $p(x|z)$ of inputs that could generate it concentrates around its maximum, we have:
    \begin{equation}
    \label{eq:shannon-fisher}
        I_{\text{eff},\beta}(x;z) \approx H(x) - \E_x\Big[\frac{1}{2} \log\Big(\frac{(2\pi e)^k}{|F_{z|x}|}\Big)\Big].
    \end{equation}
    Hence, by (i), when the Fisher of the weights decreases, the effective mutual information between inputs and activations also decreases.
\end{proposition}
\cut{Thus, decreasing the Fisher Information that the weights contain about the training set (which can be done by controlling the noise of the stochastic optimization scheme) decreases the apparently unrelated effective information between inputs and activations at test time.} Moreover, making $\nabla_x f_w(x)$ small, \ie, reducing the \textit{Lipschitz constant} of the network, also reduces the effective information.
Proposition 3.1 of \citet{achille2018emergence} states that a representation $z$ is maximally invariant to all nuisances at the same time if and only if $I(x, z)$ is minimal among sufficient representations. Together with \Cref{prop:re-emergence}, this allows us to conclude that a network with minimal information in the weights is forced to learn a representation that is \emph{effectively invariant} to nuisances.
\cut{That is, invariance emerges naturally during training by reducing the amount of information stored in the weights.} \cut{Note that there may still be information on nuisances in the weights, but it is not accessible from the network.}

\begin{figure*}
    \centering
    \includegraphics[width=.9\linewidth]{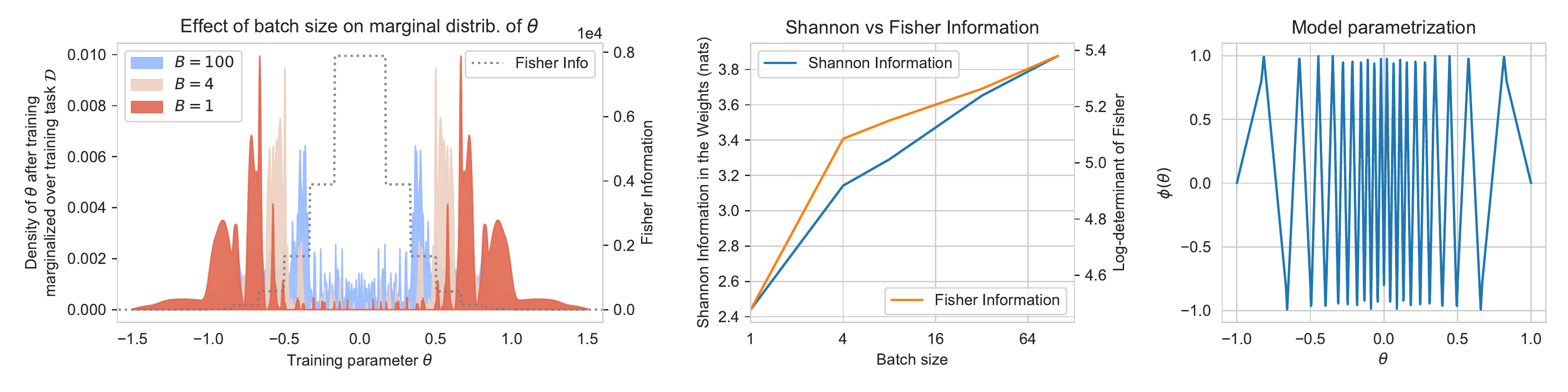}
    \caption{\textbf{Relation between curvature, Fisher and Shannon Information.}  \textbf{(Left)} Plot of the marginal distribution of the parameter $w$ at the end of training, marginalized over all possible training tasks $\mathcal{\D}$, as the batch size $B$ of SGD changes. As the batch size gets smaller, SGD shifts farther away from areas with high Fisher Information (dotted line), supporting \Cref{prop:kramers}.
    \textbf{(Center)} Effect of the batch size on the Information in the Weights. Remarkably, while changing the batch size should only affect the Fisher Information, it also reduces the Shannon Information of the weights following the same qualitative dependence, in support of \Cref{prop:weight-shannon-fisher-relation}.
    \textbf{(Right)} Redundant parametrization $\phi(\theta)$ used in the experiment to emulate some of the key properties of the loss landscape of deep networks.
    }
    \label{fig:marginal-inforamtion}
\end{figure*}

\section{Empirical validation}
\label{sec:empirical-verification}

\paragraph{Relation between curvature, Fisher and Shannon Information.}
In this section, we want to empirically verify the link between Fisher Information and Shannon Information in a model trained with SGD, as explored in \Cref{sec:sgd-dynamics}.
Our main objective is to verify that decreasing the Fisher Information (which can be done by changing the hyper-parameters of SGD, in particular the batch size and learning rate, see \Cref{prop:kramers}), does indeed decrease the Shannon Information (\Cref{fig:marginal-inforamtion}, center).
It is known in general that the Fisher Information can be used to upper-bound the Shannon Information (recall from \Cref{sec:complexity-measures} that the Shannon Information is the minimum attainable).
However, we show empirically here that even for a simple 1D example this bound is remarkably loose (by orders of magnitude).
A stronger connection between Fisher and Shannon can be obtained from \Cref{prop:weight-shannon-fisher-relation}.

In general, computing the Shannon Information $I(\D; w)$ between a dataset $\D$ and the parameters $w$ of a model is not tractable. However, here we show an example of a simple model that can be trained with SGD, replicates some of the aspects typical of the loss landscape of DNNs, and for which both Shannon and Fisher Information can be estimated easily.
According to our predictions in \Cref{sec:sgd-dynamics},  \Cref{fig:marginal-inforamtion} shows that (center) increasing the temperature of SGD, for example by reducing the batch sizes, reduces both the Fisher and the Shannon Information of the weights (\Cref{prop:weight-shannon-fisher-relation}), and (left) this is due to the solution discovered by SGD concentrating in areas of low Fisher Information of the loss landscape when the temperature is increased (\Cref{prop:kramers}).

\begin{figure*}[t]
    \centering
    \includegraphics[width=.6\linewidth]{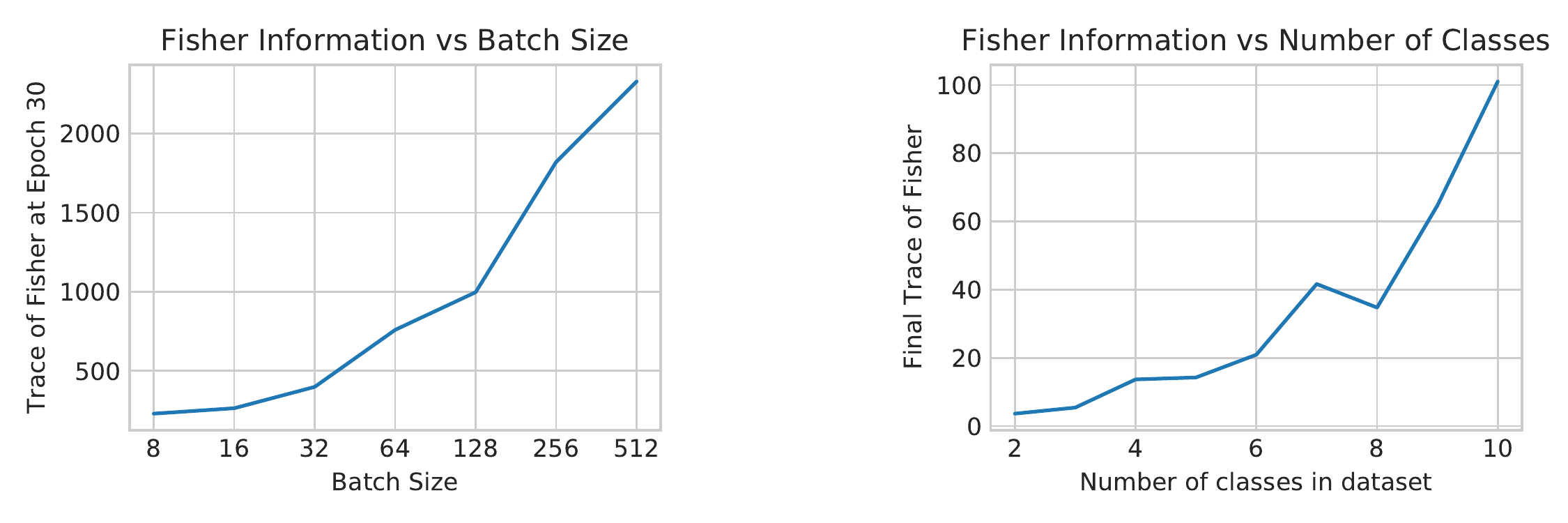}
    \caption{
    \textbf{Fisher Information in CIFAR-10.}
    \textbf{(Left)} Trace of the Fisher Information in the weights of a ResNet-18, after 30 epochs of training on CIFAR-10, for different values of the batch size.
    A smaller batch size yields a smaller Fisher Information, supporting \Cref{prop:kramers}.
    \textbf{(Right)} Fisher Information at the end of the training on a subset of CIFAR-10 with only the first $k$ classes: After training on fewer classes, the network has less Information in the Weights, suggesting that the Information in the Weight has a semantic role.
    }
    \label{fig:cifar-10}
\end{figure*}

The toy model is implemented as follow.
The dataset $\D = \{x_i\}_{i=1}^N$, with $N=100$, is generated by sampling a mean $\mu \sim \operatorname{Unif}[-1,1]$ and sampling $x_i \sim N(\mu, 1)$. The task is to regress the mean of the dataset by minimizing the loss $L_\D = \frac{1}{N} \sum_{i=1}^N (x_i - \phi(\theta))^2$, where $\theta$ are the model parameters (weights) and $\phi$ is some fixed parametrization. To simulate the over-parametrization and complex loss landscape of a DNN, we pick $\phi(\theta)$ as in \Cref{fig:marginal-inforamtion} (right).
Notice in particular that multiple value of $\theta$ will give the same $\phi(\theta)$: This ensures that the loss function has many equivalent minima. However, these minima will have different sharpness, and hence Fisher Information, due to $\phi(\theta)$ being more sharp near the origin.
\Cref{prop:kramers} suggests that SGD  is  more likely to converge to the minima with low Fisher Information.
This is confirmed in \Cref{fig:marginal-inforamtion} (left), which shows the marginal end point over all datasets $\D$ and SGD trainings.
Having found the marginal $Q(w)$ over all training and datasets, we can compute the the Shannon Information $I(\D; \theta) = \E_\D[\KL{Q(w|\D)}{Q(w)}]$.
Note that we take $Q(w|\D) = N(w^*, F^{-1})$, where $w^*$ is the minimum recovered by SGD at the end of training and $F^{-1}$ is minimum variance of the estimation, which is given by the Cram\'er-Rao bound. The (log-determinant of) Fisher-Information can instead easily be computed in closed form given the loss function $L_\D$. In \Cref{fig:marginal-inforamtion} (center) we show how these quantities change as the batch size $B$ varies.  Notice that when $B=N=100$, the algorithm reduces to standard gradient descent, which maintains the largest information in the weights.

The value of the Fisher Information cannot be directly compared to the Shannon Information, since it is defined modulo an additive constant due to the improper prior. However, using a proper Gaussian prior that leads to the lowest expected value, we obtain a value of the ``Gaussian'' Information in the Weights between 4000-5000 nats, versus the $\sim\! 4$ nats of the Shannon Information:  minimizing a much larger (Fisher) bound, SGD can still implicitly minimize the optimal Shannon bound.

\paragraph{Stability, Fisher and Information.}
In \Cref{fig:stability}, we show in a simple case the relation between stability of the training SGD path to perturbations of the dataset (the term $\nabla_\D w^*$ in \Cref{prop:weight-shannon-fisher-relation}) and the geometry of the loss landscape (the curvature, approximated by $F(w)$ at convergence).
The plot was obtained by pre-training an All-CNN architecture \cite{springenberg2014striving} on the MNIST digits 2 to 9. We pre-train the network for 10 epochs, and we fix the final point as the network initialization. We then train with SGD (learning rate 0.01, momentum 0.9, weight decay 0.0005) on the binary classification dataset $\D$ obtained by restricting to the previously unseen digits 0 and 1.
This gives us the first training path, shown in green. We then slightly perturb the dataset $\D$ by replacing a single sample (picked at random) with another sample in the dataset. This gives us a perturbed dataset $\D'$. We then train with the same settings and same random seed on this dataset, which gives us a slightly perturbed training curve (shown in red). The difference between the final point of the two curves is an approximation of $\nabla_\D w^*$ and is related to the notion of stability of SGD \cite{hardt2015train}.
To plot the curves, we project the high-dimensional weights to the 2D plane connecting the initial point and the end points of the two curves.
We then compute the diagonal of the Fisher Information Matrix, project it to this plane obtaining a 2x2 matrix, and plot its inverse as an ellipse. Thus, the axes of the ellipse show the curvature of the loss function along the plane (large axes means low curvature). In this context, \Cref{prop:weight-shannon-fisher-relation} can be interpreted as saying that if the perturbation of the final point due to any perturbation of the dataset is small compared to the curvature of the loss at that point, then the amount of information contained in the weights is also small.

\paragraph{Fisher Information and dynamics of feature learning.}

We now investigate how the amount of information in the weights of a deep neural network changes during training, in particular to see whether changes in the Fisher Information correspond to the network learning features of increasing complexity.
In \Cref{fig:bottleneck} we train a 3-layer fully connected network on a simple classification problem of 2D points and plot both the Fisher and the classification boundaries during training. Since the network is relatively small, in this experiment we compute the Fisher matrix directly from the definition.
As different features are learned, we observe corresponding ``bumps'' in the Fisher Information matrix.
This is compatible with the hypothesis advanced by \citet{achille2018critical}, whereby feature learning may correspond to crossing a narrow bottleneck (high curvature, and high Fisher) in the loss landscape, followed by a compression phase as the network moves away toward flatter area of the loss landscape. Note that ``phase transitions'' in feature learning -- where a feature is suddenly learned, following a period in which the representation is relatively stable -- are also observed when analyzing activations, rather than the weights, from an information theoretic point of view \citep{burgess2018understanding,rezende2018taming}. The concordance of the two analyses brings supports to the connection we develop in \Cref{prop:re-emergence}.

\paragraph{Fisher Information for CIFAR-10.}

In this section, we show the trade-off between amount of Information in the Weights, complexity of the task (\Cref{fig:cifar-10}, right), and value of $\beta$ (\Cref{fig:cifar-10}, left) on a more realistic problem. More precisely, we validate our predictions on a larger scale off-the-shelf ResNet-18 trained on CIFAR-10 with SGD (with momentum 0.9, weight decay 0.0005, and annealing the learning rate by 0.97 per epoch).

First, we compute the Fisher Information (more precisely, its trace) at the end of training for different values of the batch size (and hence of the ``temperature'' of SGD). In accordance with \Cref{prop:kramers}, \Cref{fig:cifar-10} (left)  shows that after 30 epochs of training the networks with low batch size have a much lower Fisher Information.
Second, to check whether the Fisher Information correlates with the amount of information contained in the dataset, we train only on some of the classes of CIFAR-10. Intuitively, the dataset with only 2 classes should contain less information than the dataset with 10 classes, and correspondingly the Fisher Information in the Weights should be smaller. We confirm this prediction in \Cref{fig:cifar-10} (right).

\section{Discussion}
\label{sec:discussion}

Once trained, deep neural networks are deterministic functions of their input, and we are interested in understanding what ``information'' they retain, what they discard, and how they process unseen data. Ideally, we would like them to process future data by retaining all that matters for the task (sufficiency) and discarding all that does not (nuisance variability), leading to invariance. But we do not have access to the test data, and the literature does not provide a rigorous or even formal connection between properties of the training set and invariance to nuisance variability in the test data.

We put the emphasis on the distinction between Information in the Weights, as done by \citet{achille2018emergence}, and information in the activations, which several other information-theoretic approaches to Deep Learning have focused on. One pertains to representations of past data, which we can measure. The other pertains to desirable properties of future data, that we cannot measure, but we can bound. We provide a  bound, exploiting the Fisher Information, which enables reasoning about ``effective stochasticity'' even if a network is a deterministic function.

Our results connect to generalization bounds through PAC-Bayes, and account for the finite nature of the training set, unlike several other information-theoretic approaches to Deep Learning that only provide results in expectation.  Also related to our work is \citet{goldfeld2018estimating}, who estimate mutual information under the hypothesis of inputs with isotropic noise. Both \citet{achille2018emergence} and \citet{shwartz2017opening} suggest connecting the noise in the weights and/or activations with the noise of SGD, although no formal connection has been established thus far.

\cut{Our results connect to generalization bounds through PAC-Bayes, and account for the finite nature of the training set, unlike several other information-theoretic approaches to Deep Learning that only provide results in expectation.} Our goal aims to explain the relations between known concepts such as {\em information}, {\em invariance}, and {\em generalization} in the context of training deep neural networks.

\section*{Acknowledgments}

Research supported in part by ARO W911NF-17-1-0304 and ONR N00014-19-1-2229.

\bibliographystyle{icml2020}
\bibliography{bibliography}

\cleardoublepage

\appendix

\section{Proofs}

\begin{proof}[\textbf{Proof of \Cref{prop:shannon-complexity-equivalence}}]
	For a fixed training algorithm $A: \mathcal{D} \mapsto  Q(w|\D)$, we want to find the prior $P^*(w)$ that minimizes the expected complexity of the data:
	\begin{align*}
	P^*(w)
	&= \argmin_{P(w)} \E_\D[C(\D)] \\
	&= \argmin_{P(w)} \Big[\E_\D[L_\D(p_w)] \\
	&+ \E_\D[\KL{Q(w|\D)}{P(w)}] \Big]
	\end{align*}
	Notice that only the second term depends on $P(w)$. Let $Q(w) = \E_\D[Q(w|\D)]$ be the marginal distribution of $w$, averaged over all possible training datasets. We have
	\begin{align*}
	\E_\D[\KL{Q(w|\D)}{P(w)}] &= \E_\D[\KL{Q(w|\D)}{Q(w)}] \\
	&+ \E_\D[\KL{Q(w)}{P(w)}].
	\end{align*}
	Since the KL divergence is always positive, the optimal ``adapted'' prior is given by $P^*(w) = Q(w)$, i.e.\ the marginal distribution of $w$ over all datasets.
	Finally, by definition of Shannon's mutual information, we get
	\begin{align*}
	    I(w; \D) &= \KL{Q(w|\D) \, \pi(\D)}{Q(w) \,\pi(\D)} \\
	    &= \E_{\D \sim \pi(\D)}[\KL{Q(w|\D)}{Q(w)}]. \qedhere
	\end{align*}
\end{proof}

\begin{proof}[\textbf{Proof of \Cref{prop:fisher-complexity-equivalence}}]

Since both $P(w)$ and $Q(w|\D)$ are Gaussian distributions, the KL divergence can be written as
\begin{align*}
\KL{Q}{P} &= \frac{1}{2} \Bigg[\frac{\norm{w^*}^2}{\lambda^2} + \frac{1}{\lambda^2} \tr(\Sigma) \\
& + k \log{\lambda^2} - \log|\Sigma| - k \Bigg],
\end{align*}
where $k$ is the number of components of $w$.
Since $\Sigma \propto \beta$, for a small $\beta$ the gaussian distribution $Q(w|\D)$ is concentrated in a sufficiently small neighbourhood of $w^*$, in which case a quadratic approximation of $L_\D$ holds. Thus, expanding $L_\D(w)$ to the second order around $w^*$ we obtain:
\begin{align*}
C_\beta(\D;P,Q) &\simeq L_\D(p_{w^*}) + \tr(H \Sigma) \\
&+ \frac{\beta}{2} \Bigg[ \frac{\norm{w^*}^2}{\lambda^2}+ \frac{1}{\lambda^2} \tr(\Sigma) + k \log \lambda^2 \\
&- \log |\Sigma|-k \Bigg].
\end{align*}
The gradient with respect to $\Sigma$ is
\[
\frac{\partial C_\beta(\D;P,Q)}{\partial \Sigma} = \bra{H + \frac{\beta}{2 \lambda^2}I -\frac{\beta}{2} \Sigma^{-1}}^\top.
\]
Setting it to zero, we obtain the minimizer $\Sigma^* =\frac{\beta}{2} (H + \frac{\beta}{2\lambda^2} I)^{-1} $.

We obtain the second part by substituting $\Sigma = \Sigma^*$ in the expression for $\KL{Q}{P}$, and the third part by applying \Cref{lemma:hessian-fisher}.
\end{proof}

\begin{proof}[\textbf{Proof of \Cref{prop:weight-shannon-fisher-relation}}]
It is shown in \citet{brunel1998mutual} that, given two random variables $x$ and $y$, and assuming that $p(x|y)$ is concentrated around its MAP, the following approximation holds:
\begin{equation}
\label{eq:fisher-shannon-approx}
I(x; y) \approx H(x) - \E_x\Bigg[\frac{1}{2} \log\Bigg(\frac{(2\pi e)^k}{|F_{y|x}|}\Bigg)\Bigg],
\end{equation}
where $F_{y|x} = \E_{y\sim p(y|x)}[-\nabla^2_x \log p(y|x)]$ is the Fisher Information that $x$ has about $y$, and $k$ is the number of components of $x$.
We want to apply this approximation to $I(w; \D)$, using the distribution $p(w|\D) = N(w^*(\D), F(w^*(\D))^{-1})$. Hence, we need to compute the Fisher Information $F_{w|\D}$ that the dataset has about the weights. Recall that, for a normal distribution $N(\mu(\theta), \Sigma(\theta))$, the Fisher Information is given by \cite{malago2015information}:
\[
F_{m,n} = \partial_{\theta_m} \mu^t \Sigma^{-1} \partial_{\theta_n} \mu + \frac{1}{2} \tr \big(\Sigma^{-1} (\partial_{\theta_m} \Sigma) \Sigma^{-1} (\partial_{\theta_n} \Sigma)\big).
\]
Using this expression in our case, and noticing that by our assumptions we can ignore the second part, we obtain:
\[
F_{w|\D} = {\nabla_\D w^*}^t F_w(w^*) \nabla_\D w^*,
\]
which we can insert in \cref{eq:fisher-shannon-approx} to obtain:
\begin{align*}
I(w; \D) &\approx H(\D) - \E_\D\Big[\frac{1}{2} \log\Big(\frac{(2\pi e)^k}{|F_{w|\D}|}\Big)\Big] \\
&= H(\D) - \E_\D\Big[\frac{1}{2} \log\Big(\frac{(2\pi e)^k}{|{\nabla_\D w^*}^t F_w(w^*) \nabla_\D w^*|}\Big)\Big].
\end{align*}
\end{proof}

\begin{proof}[\textbf{Proof of \Cref{prop:re-emergence}}]
(1) We need to compute the Fisher Information between $z_n$ and $x$, that is:
\[
F_{z|x} = \E_{z \sim p(z|x)}\Big[ - \nabla^2_x \log p(z=f_w(x)|x)  \Big].
\]
In the limit of small $\beta$, and hence small $n$, we expand
$z_n=f_{w+n}(x)$ to the first order about $w$ as follows:
\begin{align*}
    z_n &= f_{w+n}(x) + J_f \cdot n  + o(\|n\|)
\end{align*}
where $J_f$ is the Jacobian of $f_w(x)$ seen as a function of $w$, with $\dim(J_f) = \dim z \times \dim w $.
Hence, given that $n \sim N(0, \Sigma_w^*)$, we obtain that $z$ given $x$ approximately follows the distribution $p(z_n|x) \sim N(f_w(x), J_f \Sigma_w^* J_f^t)$.

We can now plug this into the expression for $F_{z|x}$ and compute:
\begin{align*}
    F_{z|x} &= \E_{z \sim p(z|x)}\big[ - \nabla^2_x \log p(z_n|x) \big]\\
    &= \frac{1}{2} \E_{z \sim p(z|x)}\Big[
    \nabla^2_x  \big[(z-m(x))^t \Sigma(x)^{-1} (z-m(x)) \\
    & + \log |\Sigma(x)|\big]\Big]\\
    &= \nabla_x f_w \cdot \Sigma_w^{-1} \,\nabla_x f_w.
\end{align*}

(2) We now proceed to estimate the Shannon mutual information $I(z;x)$ between activations and inputs. In general, this does not have a closed form solution, rather we use again the approximation of \citet{brunel1998mutual} as done in the proof of \Cref{prop:weight-shannon-fisher-relation}. Doing so we obtain:
\begin{align*}
    I_\text{eff}(x;z) &\approx H(x) - \E_x\Big[\frac{1}{2} \log\Big(\frac{(2\pi e)^k}{|F_{z|x}|}\Big)\Big] \\
    &\approx H(x) - \E_x\Big[\frac{1}{2} \log\Big(\frac{(2\pi e T)^k}{|\nabla_x f_w \cdot J_f F_\w J_f^t\,\nabla_x f_w|}\Big)\Big].
\end{align*}
\end{proof}

\end{document}